\pdfoutput=1

\documentclass[11pt]{article}

\usepackage{EMNLP2023}

\usepackage{times}
\usepackage{latexsym}
\usepackage{tikz}
\usetikzlibrary{decorations.pathreplacing,positioning,shadows}
\usetikzlibrary{trees}
\usepackage{lipsum}
\usepackage{booktabs}
\usepackage{graphicx}
\usepackage{multirow}
\usepackage{listings} 
\usepackage{xcolor} 
\usepackage{pifont}
\usepackage[T1]{fontenc}
\usepackage{color,soul}
\usepackage[utf8]{inputenc}
\usepackage{enumitem}

\definecolor{cadmiumgreen}{rgb}{0.0, 0.42, 0.24}
\definecolor{celestialblue}{rgb}{0.29, 0.59, 0.82}

\newcommand{\myblue}[1]{\textcolor{celestialblue}{#1}}
\usepackage{microtype}

\usepackage{inconsolata}

\title{NarrativePlay: Interactive Narrative Understanding}

\author{Runcong Zhao$^{1}$, Wenjia Zhang$^2$,  Jiazheng Li$^{1}$, Lixing Zhu$^{1}$, \\ 
\textbf{Yanran Li, Yulan He$^{1,2,3}$, Lin Gui$^1$}\\
  $^1$King's College London, $^2$University of Warwick, $^3$The Alan Turing Institute\\
  \texttt{\{runcong.zhao, wenjia.1.zhang, jiazheng.li, lixing.zhu\}@kcl.ac.uk} \\ \texttt{yanranli.summer@gmail.com}, \texttt{\{yulan.he, lin.1.gui\}@kcl.ac.uk} }

\begin{document}
\maketitle
\begin{abstract}
In this paper, we introduce NarrativePlay, a novel system that allows users to role-play a fictional character and interact with other characters in narratives such as novels in an immersive environment. 
We leverage Large Language Models (LLMs) to generate human-like responses, guided by personality traits extracted from narratives. The system incorporates auto-generated visual display of narrative settings, character portraits, and character speech, greatly enhancing user experience. Our approach eschews predefined sandboxes, focusing instead on main storyline events extracted from narratives from the perspective of a user-selected character. NarrativePlay has been evaluated on two types of narratives, detective and adventure stories, where users can either explore the world or improve their favorability with the narrative characters through conversations.
\end{abstract}

\section{Introduction}
People's experiences and thought processes can be effectively stored in a database, serving as a valuable repository of personality traits. Recent studies \citep{generative-agent2023, autogpt2023, Ouyang2022TrainingLM} have leveraged Large Language Models (LLMs) to generate human-like responses, which are guided by relevant memories retrieved from such a personality database when prompting LLMs. This significant advancement presents an exciting opportunity for creating an immersive and interactive environment that could enable emulating the dynamic storylines one might encounter while reading books, akin to those featured in the television series ``Westworld''. 

However, existing approaches which leveraged LLMs for the generation of interactive agents often assume predefined personality traits. For instance, \citet{generative-agent2023} used a short narrative to seed each agent's identity, while \citet{npc2023} tailored non-player character (NPC) characteristics according to game-relevant features. Narratives contain extensive character-centric details such as ``Who'', ``When'', ``Where'', ``Action'', ``Objective'', and ``Appearance'', which can be utilised to craft vivid characters. While extracting comprehensive character traits from long and complex narratives is challenging and remain largely under-explored \citep{xu-etal-2022-fantastic}, we show in this paper how to leverage the strong zero-shot learning capability of LLMs to create interactive agents. 

\begin{figure*}[t]
    \centering
    \includegraphics[width=1\linewidth]{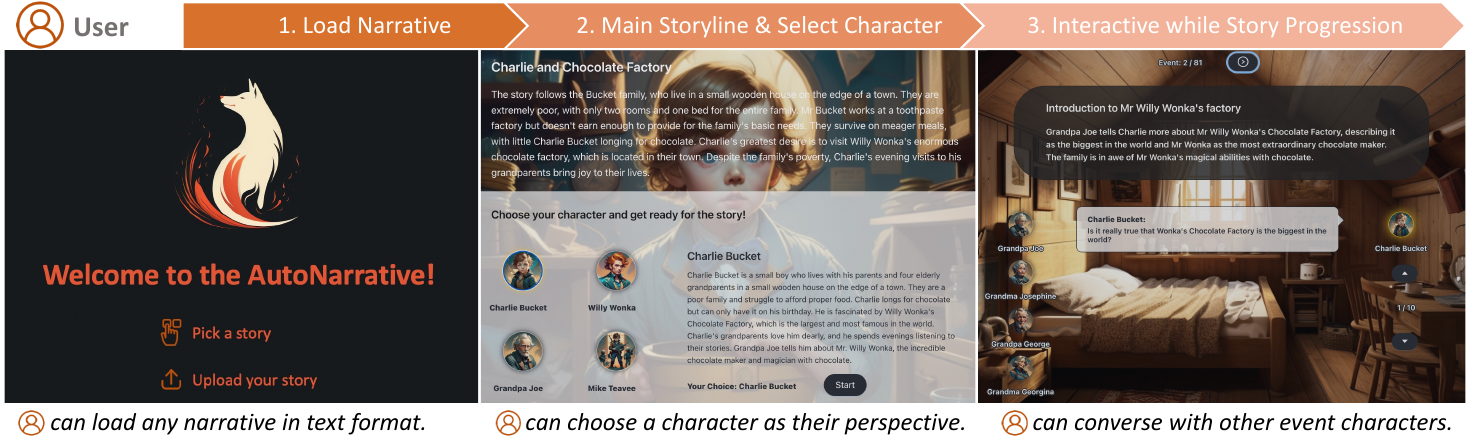}
    \caption{Our system's interactive process begins when a user provides a  narrative 
    to the system. They then choose a character as their narrative identity, through whom they can engage with the story. Users can have conversations with other characters, thereby experiencing the story in a more immersive way.}
    \label{fig:progression}
\end{figure*}

Creating interactive and immersive environments for users and agents can be challenging due to two key factors: (1) \emph{Environment Extraction}. Environments or narrative settings are often vaguely defined unless crucial to the plot. Existing research predominantly concentrates on agent behaviours within manually constructed sandboxes \citep{Riedl_Bulitko_2012, cote18textworld, jericho2020, generative-agent2023}, which requires significant manual efforts and lacks generalisability. We propose an approach focusing on main storyline events from the perspective of a user-selected character, reducing the complexity of identifying narrative settings. (2) \emph{Environment Generation}. Leveraging stable diffusion models \cite{9423212, 9878449} as external knowledge \citep{flamingo2022}, we use image generation models to fill in missing details in environments. While generating knowledge for specific narrative settings is challenging, models trained on certain image styles, like fairy tales or animations, excel in this task.

We categorise user (or player) behaviours and compile commonly asked questions in interactive narratives to evaluate the quality of the agents' responses. As we design interactive narratives in a novel setting compared to previous works, we have developed approaches which address certain limitations of existing works. 
Our developed NarrativePlay opens up an interesting avenue of 
interactive narrative understanding. 

In summary, we have made the following contributions:

\begin{itemize}[noitemsep,nolistsep]
    \item We have developed NarrativePlay, a novel web-based platform capable of transforming narrative inputs into immersive interactive experiences. Our system synchronises text with visual displays of story settings, character portraits and speech, leveraging advanced multi-modal LLMs to enhance user experience.
    \item We have proposed to extract comprehensive character traits from narratives for authentic characters that generate human-like responses and adhere to predefined roles.
    \item Instead of using resource-intensive and less versatile predefined sandboxes, our approach focuses on main storyline events from narratives. We simplify the complex world into visuals from a user-chosen perspective, enhancing adaptability.
    \item We have overcome the challenges of generating environments with reasonable details by leveraging stable diffusion models as an external knowledge base.
    \item We have categorised player behaviours and compiled common questions in interactive narratives to assess the quality of agent responses. 
\end{itemize}

\section{Architecture of NarrativePlay}
\begin{figure*}[h]
    \centering
    \includegraphics[width=\linewidth]{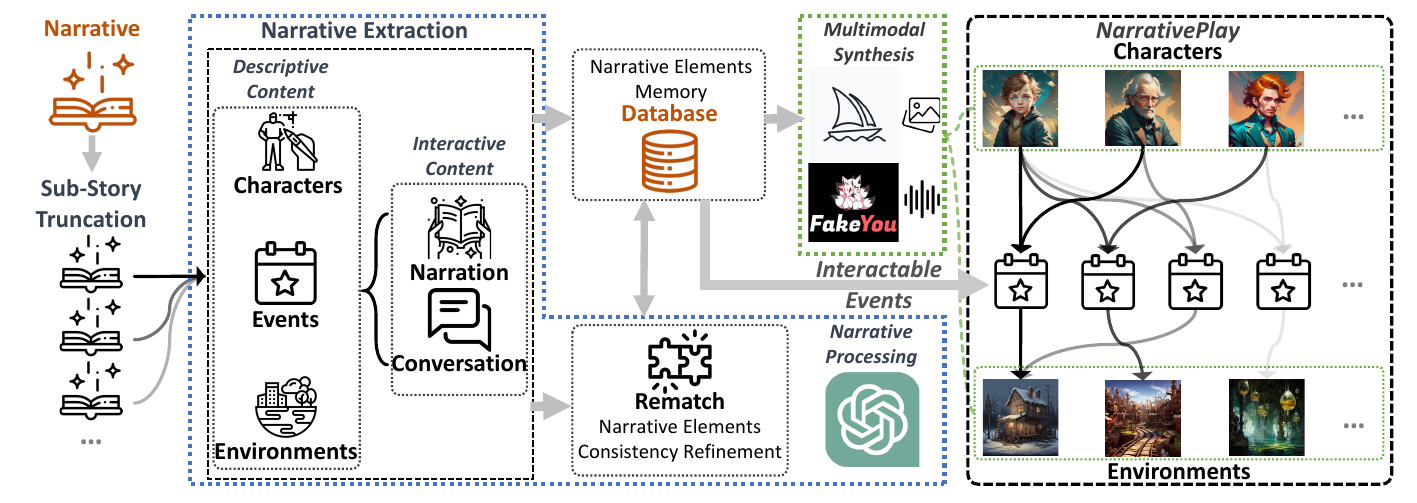}
    \caption{Demonstration of our \textbf{NarrativePlay} through a pipeline view.}
    \label{fig:architecture}
\end{figure*}

Figure~\ref{fig:architecture} shows an overview of NarrativePlay, including three modules: (1) main storyline extraction; (2) narrative image and speech synthesis; and (3) main storyline progression. 

\subsection{Main Storyline Extraction} \label{sec:main-storyline-extraction}
We utilise the OpenAI API to extract structured information from text, including characters, events, conversations, and environments, employing the most recent ChatGPT model \texttt{gpt-3.5-turbo}. 
Because of the input limitation, we first divide an input narrative into smaller chunks, ensuring that individual sentences are kept intact. 

A notable challenge when utilising LLMs for Information Extraction (IE) is that the LLM-generated outputs do not always follow our desirable format, as shown in \textsection{\ref{sec:prompt-format-error}}. In our context, we employ the JSON formatting, system prompts, and example outputs to mitigate such errors. Additionally, we implement post-processing to address issues such as missing punctuation, and, if errors persist, we instruct the GPT model to rectify them. Despite these attempts, we can only reduce, not completely eliminate, the output formatting errors. In scenarios where the output format remains incorrect, NarrativePlay outputs either null results or some predefined values, depending on whether there will be a potential disruption to the narrative flow. For instance, if the objective of a character can not be extracted, the system will leave this field empty. 
Conversely, for character appearance, we randomly choose a hair colour and an eye colour as the default. Leaving the character appearance field blank may lead to inconsistent attributes, for example, the character's eye colour may change from blue to brown in the middle of a narrative. 


In what follows, we describe in detail how we extract various narrative elements using ChatGPT.

\paragraph{Character} For an input narrative $S$, our initial step is to solicit a list of the characters involved. We highlighted the prompt and variable(s) in blue in our designed prompt template below:
\begin{lstlisting}[mathescape=true]
(*@\myblue{Find all characters in the given story, return in}@*)
(*@\myblue{JSON format}@*) 
Extract characters in the story, here is the format example: [{(*@``@*)name(*@''@*): (*@``@*)Charlie Bucket(*@''@*)}, {(*@``@*)name(*@''@*): (*@``@*)Grandpa Joe(*@''@*)}]
Here is the story: (*@\myblue{$S$}@*) 
\end{lstlisting}
Subsequently, for each newly occurred character $c$, we additionally extract their defining characteristics. This includes their core traits, appearance, and quotes. These elements are extracted separately because we have observed that the GPT model tends to introduce more formatting errors when tasked with extracting a larger set of defining characteristics at once. Prompts used can be found in \textsection{\ref{sec:char-prompt}}.

\paragraph{Events}
For each event, we extract the description, the characters involved, the location, and the conversation that takes place during the event using prompt in \textsection{\ref{sec:event-prompt}}. This approach allows us to link each event with its corresponding characters and locations, thereby eliminating the need to extract the timeline of the story. This strategy is particularly useful for narratives where multiple events could be described simultaneously, making it difficult to disentangle them. We assume that a character cannot be present in multiple events simultaneously. Though this assumption may not hold all the time, we found that it works well in our evaluated scenarios. We leave the extraction of more complex events as our future work. 

If multiple characters are involved in an event, 
we will also attempt to extract conversation in the event using the prompt in \textsection{\ref{sec:conv-prompt}}. We extract conversations for two reasons: firstly, there is no need to further extract the embedded subevents (if any) as they are captured in conversational content. 
Secondly, it allows for a smoother transition to new conversations between users (i.e., users' chosen narrative characters) and agents (i.e., other characters in a narrative).

\paragraph{Environments}
Existing research predominantly concentrates on agent behaviours within manually constructed sandboxes, where environments, agents, and actions  
 are pre-defined \citep{cote18textworld, jericho2020}.  However, constructing sandboxes for narratives with diverse backgrounds and settings is resource-intensive and lacks generalisability. 

Character locations and environments, unless vital to the plot, are often vaguely described in narratives and may thus require clarification, which makes automatic extraction very challenging. To overcome this, we propose focusing on location of main storyline events extracted from narratives and creating environment visuals rooted in the event descriptions. Prompts used can be found in \textsection{\ref{sec:env-prompt}}. Similar to event extraction, we used example outputs obtained from the story using GPT-4 to boost performance.

It is common to observe multiple mentions of the same location in narratives. 
For example, 
``Old people's room'', ``Grandparents' room'' and ``Bedroom'' all refer to the same place. Additionally, vague descriptions such as ``Various locations'', ``Their house'', and ``Unknown'' are common and further complicate the environment extraction task. Generating images from event environment descriptions partly alleviates the issues of location co-referencing. 
Moreover, while capturing dynamic changes in location attributes, like the onset of snowfall in winter, is challenging when extracted directly from narratives, such details can be more easily 
represented in the generated images. For cases where the detailed description of environments is missing in narratives, we will discuss in the next subsection how to synthesise images which fill in the missing details using stable diffusion models.

While fostering meaningful interactions among users and agents without traditional sandbox constraints is challenging, our solution reduces the complexities of the world from the user-selected character's perspective. We guide the visibility among agents via shared event participation.

\subsection{Narrative Image and Speech Synthesis}
\paragraph{Narrative Image Synthesis} 

We leverage the stable diffusion models as external knowledge \citep{flamingo2022} to generate scenarios in situations where information is lacking. While creating specialised knowledge bases for specific narrative worldviews (e.g., magical realms, post-apocalyptic wastelands, futuristic settings) remains a challenge, we adapt models trained on specific picture styles, such as fairy tales and oil painting, to auto-complete the intricate details of the environment. 

We utilise character and event features extracted for the text-to-image generative models as we discussed above. Our framework offers two modes of image synthesis: (1) \textbf{Local Synthesis}: For users with substantial compute resources, an open-source text-to-image model, \texttt{openjourney}, accessible via HuggingFace\footnote{\url{https://huggingface.co/prompthero/openjourney}}, is used to generate images locally. (2) \textbf{Cloud-based Synthesis}: For users with limited compute resources, we have incorporated an API request-based image generation service offered by Hotpot AI\footnote{\url{https://hotpot.ai/}} into our framework for 
generating character portraits, which offers a more stable generation style. Additionally, for event image generation, we employ Midjourney\footnote{\url{https://www.midjourney.com/}} as it provides more varieties and detailed pictures.

While advancements in video synthesis have been notable \cite{singer2022makeavideo}, the considerable computational resources required, coupled with the subpar quality of the generated video, presently render the user experience suboptimal, thus precluding its implementation at this stage.

\paragraph{Narrative Speech Synthesis} Our multimodal synthesis framework also includes the transformation of narrative text into compelling speech, enriching the experience with an auditory dimension. For this crucial task, we primarily employ Text-to-Speech (TTS) models from the FakeYou\footnote{\url{https://fakeyou.com/}} platform
, which offers over three thousand models, allowing each narrative character a unique voice. With the extensive TTS model assortment from FakeYou, our framework facilitates the creation of diverse and captivating narrative experiences.

A noteworthy feature of our approach is real-time text-to-speech conversion, creating an interactive and immersive storytelling environment that sustains user engagement.


\subsection{Main Storyline Progression}
As shown in Figure~\ref{fig:progression}, we progress the main storyline with three stages:
\paragraph{Narrative Input} In the process of creating an interactive narrative with our system, the user begins by selecting or uploading their chosen narrative. 

\paragraph{Character Selection} Following above, NarrativePlay extracts the main storyline and subsequently presents the information about the background and characters. Users are then asked to select from the listed major characters to begin their adventure  \cite{dominguez2016mimesis}, which are defined as those involved in at least $20\%$ of the events. We restrict users' choices to the top major characters in order to have a better story flow. 
The agent's memory is initialised at this stage, laying the groundwork for future interactions. 

\paragraph{Story Progression} Once a character is selected, we present events related to the chosen character to the user. The location image is displayed as the background picture and the event description appears as a narration in the black box at the top of the page. Each event displays the involved characters on the left, with the user-selected character on the right. If there are conversations extracted for this event, they will be played first with voice renditions. Then, the user can click on any other character to engage in conversations with them. During this stage, NarrativePlay retrieves the most relevant, recent, and important memories from the agent's past, ensuring continuity and context-awareness \cite{agencyplay2009} in the generated responses. NarrativePlay also updates the agents' memories in accordance with the progression of events, user inputs, and agent responses.

We assign a weight $w_m$ to each memory $m$ to retrieve the top memories for use in the prompt. Consequently, the weight of each memory, given the input $x$, is defined as:
$w_m = \frac{\mathbf{h_m}\cdot\mathbf{h_x}}{{\|\mathbf{h_m}\| \|\mathbf{h_x}\|}} + c^{(I-i)} + s_m$, 
where $\mathbf{h_m}$ is the embedding for memory $m$, $\mathbf{h_x}$ is the embedding for input $x$, $c$ is the decay factor set to 0.99, $i \in \{0, 1, ... I\}$ is the event index (with I being the current event index), and $s_m$ is the importance score given by GPT-3.5 based on the character and the memory. In summary, this equation denotes: \texttt{Retrieval Weight $=$ Relevance $+$ Recency $+$ Importance}.

We generate responses using the character information, the current events, the user input, and the retrieved memory using prompt in \ref{sec:conversation-prompt}. When a user selects a character to interact with, we assume the user's character is approaching the chosen character. There is a chance $p$, dependent on the relationship between the two characters, that the chosen character might initiate a conversation.



\section{Evaluation}
Evaluating such a system is challenging due to the lack of gold-standard responses, especially about events and environments. Human assessment demands deep narrative understanding, making it costly, and subjective interpretations may cause low inter-annotator agreement. 

We instead recruit three annotators to read whole narratives and rate responses to our specifically designed questions. We also explore automatic evaluation methods using LLaMA-2-70B \citep{llama22023}. We did not use GPT-4 for this purpose, as it shares a significant amount of training data with GPT-3.5, which could lead to an unfair evaluation. 
For each evaluation aspect, we provide detailed instruction, including the corresponding rubric and evaluation examples, to help both human annotator and LLaMA to understand our scoring instruction. 

Evaluations are conducted on two distinct narrative types: the adventure story \emph{Charlie and the Chocolate Factory} (CCF) and the detective novel \emph{Murder on the Orient Express} (MOE).

\subsection{Evaluation Schema}
\paragraph{Player Questions} 
We categorise player behaviours and outline questions that might be commonly asked by players in interactive narratives into the following types: (1) \emph{Character}: Questions related to the characters themselves, which could be about their background explicitly stated in the story or traits that can only be implied from the story, such as ``What is your favourite colour?''. (2) \emph{Clarification}: Questions arise when a player is confused or requires more information. They might ask for explanations of story elements, reminders of objectives, or clarifications about confusing events or instructions. This requires the capability to accurately recall specific events or dialogues from their memory. (3)  \emph{Relationships}: Queries concern the relationships between characters, such as their current status, history, or potential developments.
(4) \emph{Strategy}: Queries to seek guidance on narrative progress, requiring the agent to recall their short-term or long-term objectives. 
This type of question various depending on a particular story, such as ``Which path should I take to reach [destination] fastest?'' in an adventure novel, and ``What is the best way to approach this puzzle?'' in a detective novel.
(5) \emph{Hypothetical}: Queries explore ``what-if'' scenarios, asking how the characters might respond under different conditions or actions.

\paragraph{Evaluation Aspects} To evaluate our system's performance, we employ the controlled assessment method used by \citet{generative-agent2023} to examine the responses from each individual agent. Inspired by the previous work in chat-oriented dialogue system evaluation \citep{abc2023}, we chose the following evaluation aspects, which are important under our interactive narrative setting: \emph{Coherence}, \emph{Relevance}, \emph{Empathetic}, \emph{Commonsense}. 

\noindent Further details on evaluation can be found in \textsection{\ref{sec:evaluation-instruction}}.

\subsection{Evaluation Results}
\paragraph{Extracted Information}
We first present the information extraction results in Table~\ref{tab:extraction-evaluation}. `Incorrect' refers to the percentage of extracted characters that do not correspond to specific characters, such as ``unknown'', ``somebody'', or ``people worldwide''. Such incorrect identifications commonly appear for characters who are not central to the main plot and might be encountered briefly without a significant role. Therefore, these errors typically have a minimal impact on the main storyline. For correctly extracted characters, we assess the accuracy of their extracted summaries, objectives, appearances, and speeches, ensuring they accurately correspond to the target character. We also evaluate the percentage of duplications (e.g., ``Mrs Caroline Hubbard'', ``elderly American lady'', and ``Linda Arden''). Duplicated characters could detrimentally affect the memories, as the memories for the same character are saved as separate entities. 

Table~\ref{tab:extraction-evaluation} indicates that detective narrative poses more significant challenges. Unlike in \emph{CCF}, where characters are introduced the first time they appeared in the story, in \emph{MOE}, characters often attempt to hide their true identities, and clues are left for readers to discover. Consequently, they often begin with an appearance description from the main character's perspective, such as ``elderly American lady'', or ``a middle-aged woman dressed in black with a broad, expressionless face. German or Scandinavian''. As the story progresses, more information about the character, including their name, experiences, and objectives, is revealed. This can confuse the model, leading it to identify descriptions at different stages as separate characters. Furthermore, objectives are challenging to identify when characters first appeared in the narrative.

\begin{table}[!ht]
\begin{center}
\resizebox{\linewidth}{!}{
\begin{tabular}{ccccccc}
\toprule
\textbf{Story} & \textbf{Incorrect} $\downarrow$  & \textbf{Duplicate} $\downarrow$ & \textbf{Summary}$ \uparrow$         & \textbf{Objective}$ \uparrow$ & \textbf{Appearance} $\uparrow$ & \textbf{Voice} $\uparrow$ \\ \midrule
CCF & 0.191 &	0.211 & 0.868 &	0.816 &	0.921 &	0.868  \\
MOE & 0.272 & 0.407 & 0.898 & 0.576 & 0.915 & 0.780     \\
\bottomrule
\end{tabular}}
\caption{Extracted Information Evaluation. }
\label{tab:extraction-evaluation}
\end{center}
\end{table}

\paragraph{Agent Responses}
As shown in Table~\ref{tab:human-evaluation}, we observed that while the agent performed well in terms of \emph{relevance} and \emph{commonsense}, it fell short in \emph{coherence} and \emph{empathy} for both narratives. For instance, agents maintained a cheerful demeanour and expressed enthusiasm for travel even after a murder. 
Besides, agents often divulged everything they knew from memory, which works for stories like \emph{CCF}, but is unsuitable for detective narratives where characters may lie to serve their interests.

Equipped with memories, NarrativePlay surpasses the baseline that lacks memory. In \emph{CCF}, major characters perform better than minor ones, likely due to their more detailed narratives guiding LLMs to better understand the characters and predict their behaviours. However, in \emph{MOE}, minor characters outperform major ones. This is likely because the more complex responses required for major characters are only minimally supported by their memories, which are saved as separate entities due to the difficulty of LLMs in dealing with multiple mentions of the same character. 

While prior studies \citep{generative-agent2023, autogpt2023} have maxed out the context window at 4,096 tokens for each ChatGPT API call to enhance reasoning and prompting, we found that a longer prompt does not necessarily yield improved performance. In fact, it may potentially distract the model from focusing on the core information. Despite our efforts to automatically adjust weights of the relevant memories, 
their significance diminishes when being incorporated into the prompt. 

\begin{table}[!ht]
\begin{center}
\resizebox{\linewidth}{!}{
\begin{tabular}{ccccccccc}
\toprule
                                    & \multicolumn{2}{c}{\textbf{Coherence}} & \multicolumn{2}{c}{ \textbf{Relevance}} & \multicolumn{2}{c}{\textbf{Empathetic}} & \multicolumn{2}{c}{ \textbf{Commonsense}} \\ 
\cmidrule(lr){2-3} \cmidrule(lr){4-5} \cmidrule(lr){6-7} \cmidrule(lr){8-9}
\multirow{-2}{*}{\textbf{Category}} & w/o                & w/                & w/o                           & w/                            & w/o                & w/                 & w/o                            & w/                             \\
\midrule
\multicolumn{9}{c}{\textit{Charlie and the Chocolate Factory}}                                                                                                                                                                                               \\
\midrule
Overall                             & 0.915              & 1.085             & 1.970                         & 1.967                         & 1.252              & 1.444              & 1.000                          & 1.000                          \\
Major Characters                    & 0.900              & 1.117             & 1.972                         & 1.950                         & 1.222              & 1.483              & 1.000                          & 1.000                          \\
Minor Character                     & 0.944              & 1.022             & 1.967                         & 2.000                         & 1.311              & 1.387              & 1.000                          & 1.000                          \\
Fleiss' kappa                       & \multicolumn{2}{c}{0.486}              & \multicolumn{2}{c}{0.317}                                     & \multicolumn{2}{c}{0.337}               & \multicolumn{2}{c}{1.000}                                    \\
\midrule
\multicolumn{9}{c}{\textit{Murder on the Orient Express}}                                                                                                                                                                                                \\
\midrule
Overall                             & 1.267              & 1.400             & 2.000                         & 2.000                         & 1.157              & 1.219              & 1.000                          & 1.000                          \\
Major Characters                    & 1.011              & 1.367             & 2.000                         & 2.000                         & 1.033              & 1.222              & 1.000                          & 1.000                          \\
Minor Character                     & 1.458              & 1.425             & 2.000                         & 2.000                         & 1.250              & 1.217              & 1.000                          & 1.000                          \\
Fleiss' kappa                       & \multicolumn{2}{c}{0.404}              & \multicolumn{2}{c}{1.000}                                  & \multicolumn{2}{c}{-0.003}              & \multicolumn{2}{c}{1.000}       \\                            
\bottomrule
\end{tabular}}
\caption{Human Evaluation results. Comparing the quality of agent responses with and without the retrieved memory, as well as the difference in response quality between major and minor characters.}
\label{tab:human-evaluation}
\end{center}
\end{table}

\begin{table}[!ht]
\begin{center}
\resizebox{\linewidth}{!}{
\begin{tabular}{ccccccccc}
\toprule
                                    & \multicolumn{2}{c}{\textbf{Coherence}} & \multicolumn{2}{c}{ \textbf{Relevance}} & \multicolumn{2}{c}{\textbf{Empathetic}} & \multicolumn{2}{c}{ \textbf{Commonsense}} \\ 
\cmidrule(lr){2-3} \cmidrule(lr){4-5} \cmidrule(lr){6-7} \cmidrule(lr){8-9}
\multirow{-2}{*}{\textbf{Category}} & w/o                & w/                & w/o                           & w/                            & w/o                & w/                 & w/o                            & w/                             \\
\midrule
\multicolumn{9}{c}{\textit{Charlie and the Chocolate Factory}}                                                                                                                                                                                               \\
\midrule
Overall                             & 1.433              & 1.333             & 1.633                         & 1.547                         & 1.587              & 1.507              & 0.613                          & 0.567                          \\
Major Characters                    & 1.467              & 1.500             & 1.567                         & 1.700                         & 1.433              & 1.467              & 0.600                          & 0.617                          \\
Minor Character                     & 1.411              & 1.222             & 1.656                         & 1.444                         & 1.667              & 1.533              & 0.400                          & 0.411                          \\
\midrule
\multicolumn{9}{c}{\textit{Murder on the Orient Express}}                                                                                                                                                                                                \\
\midrule
Overall                             & 1.000              & 0.960             & 0.933                         & 0.753                         & 1.213              & 1.107              & 0.640                          & 0.640                          \\
Major Characters                    & 1.367              & 1.367             & 1.000                         & 1.100                         & 1.333              & 1.333              & 0.733                          & 0.733                          \\
Minor Character                     & 0.822              & 0.689             & 0.767                         & 0.522                         & 1.056              & 0.956              & 0.489                          & 0.489                          \\
\bottomrule
\end{tabular}}
\caption{Automatic Evaluation using LLaMa-2-70B. LLaMa evaluates responses from major characters higher than those from minor characters and rates responses without memory usage higher than those with memory. We found there are still gaps between human understanding and LLaMa. }
\label{tab:automatic-evaluation}
\end{center}
\end{table}

\section{Conclusions and Future Work}
NarrativePlay, a novel platform, transforms narratives into interactive experiences, addressing challenges of storyline extraction, authentic character creation, and versatile environment design. By focusing on the main events and leveraging advanced LLMs, it aligns text, image, and speech, marking a step forward in immersive interactive narratives. Furthermore, we categorise player behaviours and design commonly asked questions to evaluate the system's performance, and provide an evaluation framework for interactive narratives. With a potential for wider applications like game generation, NarrativePlay paves the way for future advancements in narrative understanding.

Our current work has the following limitations. First, due to the lack of an API from Midjourney, manual input of GPT-generated prompts is necessary. Although we provide HotPot API as an automatic substitution, the quality of its generated pictures is inferior to those from Midjourney. Second, the prolonged waiting time for the FakeYou API adversely affects real-time generation, slowing narrative progression and potentially impairing user experience. Controlling the quality of API-generated content is challenging. Third, we assume a linear event timeline in the input narrative. This assumption implies that there are no time jumps or flashbacks in the story. Future work needs to explore ways to deal with more complex storylines in narratives. Fourth, human evaluation is expensive. For future work, we plan to log users' activities and feedback to provide data for evaluating and further improving the performance of our system.

\section*{Ethics}
Although we have not identified any harmful outputs from ChatGPT in our study, it is worth noting that previous research has observed instances where ChatGPT produced unexpected results. We encourage other researchers to utilise this framework to scrutinise the output generated from specific prompts in ChatGPT that may have the potential to generate harmful information.


\bibliography{anthology,custom}
\bibliographystyle{acl_natbib}

\newpage
\appendix

\section{Prompt} \label{sec:prompt}

\subsection{Response Format Errors} \label{sec:prompt-format-error}

\paragraph{Unwanted Output}
GPT is trained as a chatbot, so it tends to provide an explanation before generating the required output, often leading to the inclusion of non-essential content. An example is shown below:

\begin{lstlisting}[mathescape=true]
In the given list, there are a few characters that can be considered duplicates based on certain keywords or names:
\end{lstlisting}

Significantly, adopting a JSON format mitigates this issue \citep{li2023overprompt}. 

\paragraph{Incomplete Response}
However, when all tasks are assigned to GPT in one go, the complexity often results in omissions of required content. Therefore, we divide the task into several steps, such as extracting characters, events, and environments, as discussed in Section \ref{sec:main-storyline-extraction}.

Additionally, there are cases where the model generates incomplete outputs, and this is not due to a maximum token limit. The cause remains unknown due to the opaque nature of the GPT model. In such scenario, NarrativePlay outputs either null results or some predefined values, depending on whether there will be a potential disruption to the narrative flow. 

\paragraph{Syntax Error}
When GPT is tasked with generating text with specific structures, it might not always do so correctly due to its limitations in understanding complex formatting rules. Syntax errors can cause challenges in parsing JSON. Here are some examples of common errors:

1. Missing commas. We implement post-processing to address issues such as missing syntax, and, if errors persist, we instruct the GPT model to rectify them. 
\begin{lstlisting}[mathescape=true]
$\{$(*@``@*)keywords(*@''@*): [(*@``@*)town(*@''@*), (*@``@*)chocolate factory(*@''@*), (*@``@*)small(*@''@*), (*@``@*)impoverished(*@''@*), (*@``@*)mysterious(*@''@*)](*@\hl{\textvisiblespace}@*)(*@``@*)description(*@'@*): (*@``@*)The town is a small and impoverished place with dull and dreary surroundings. Most of the residents live in humble conditions, struggling to make ends meet.(*@'@*)$\}$
\end{lstlisting}

2. Mixed double quotes and single quotes. JSON need double quote on string to be parsed. We found that, in most cases, this issue can be addressed by using examples with double quotes in the prompt, guiding GPT to adhere to our preferred format.
\begin{lstlisting}[mathescape=true]
$\{$(*@\hl{`}@*)keywords(*@\hl{'}@*): [(*@\hl{`}@*)town(*@\hl{'}@*), (*@\hl{`}@*)chocolate factory(*@\hl{'}@*), (*@\hl{`}@*)small(*@\hl{'}@*), (*@\hl{`}@*)impoverished(*@\hl{'}@*), (*@\hl{`}@*)mysterious(*@\hl{'}@*)], (*@\hl{`}@*)description(*@\hl{'}@*): (*@\hl{`}@*)The town is a small and impoverished place with dull and dreary surroundings. Most of the residents live in humble conditions, struggling to make ends meet.(*@\hl{'}@*)$\}$
\end{lstlisting}

3. Quotation marks within the value of a dictionary. A similar correction can be achieved in most cases by adding a backslash before double quotes in the prompt examples to guide GPT to use the escape character.
\begin{lstlisting}[mathescape=true]
$\{$(*@``@*)speaker(*@''@*): (*@``@*)Grandpa Joe(*@''@*), (*@``@*)content(*@''@*): (*@``@*)And then Mr Slugworth's factory began making sugar balloons that you could blow up to huge sizes before you popped them with a pin and gobbled them up. And so on, and so on. And Mr Willy Wonka tore his beard and shouted, (*@\hl{``}@*)This is terrible! I shall be ruined! There are spies everywhere!(*@''@*)$\}$
\end{lstlisting}

\subsection{Prompt for Character Extraction} \label{sec:char-prompt}
To establish the personality for each character as an interactive agent, we extract their core traits using the following prompt:
\begin{lstlisting}[mathescape=true]
(*@\myblue{Derive details pertaining to the specified}@*)
(*@\myblue{character from the provided text. In case the}@*) 
(*@\myblue{text does not contain sufficient information, }@*)
(*@\myblue{make an educated inference. Present the output in}@*)
(*@\myblue{JSON format}@*) 
Generate the character background summary, keywords, and the objective of (*@\myblue{$c$}@*). The output format is {(*@``@*)summary(*@''@*): (*@``@*)here is the background(*@''@*), (*@``@*)keywords(*@''@*): (*@``@*)personality keywords(*@''@*), (*@``@*)objective(*@''@*): (*@``@*)character's objective(*@''@*)}.
Here is the story: (*@\myblue{$S$}@*) 
\end{lstlisting}
In order to generate picture for each character, we extract their appearance using the prompt:
\begin{lstlisting}[mathescape=true]
(*@\myblue{Imagine the appearance of the specified character }@*) 
(*@\myblue{from the provided text. Present the output in}@*)  
(*@\myblue{JSON format.}@*) 
Generate the appearance, gender and age of $c$. (*@\myblue{\{character~description\}}@*) The output format is {(*@``@*)appearance(*@''@*): (*@``@*)brown hair, blue eyes, poor(*@''@*), (*@``@*)gender(*@''@*): (*@``@*)male(*@''@*), (*@``@*)age(*@''@*): (*@``@*)middle age(*@''@*)}.
\end{lstlisting}
In order to match the voice of each character, we classify their gender and age using the prompt:
\begin{lstlisting}[mathescape=true]
(*@\myblue{Identify the character's gender and age. Present}@*) 
(*@\myblue{the output in JSON format.}@*) 
Identify the gender and age of $c$. (*@\myblue{\{character~description\}}@*) For gender, choose from `male' or `female'. For age, choose from `child', `yongth', `middle age', `old age'. The output format is {(*@``@*)gender(*@''@*): (*@``@*)male(*@''@*), (*@``@*)age(*@''@*): (*@``@*)child(*@''@*)}. 
\end{lstlisting}

\subsection{Prompt for Event Extraction} \label{sec:event-prompt}
To extract events, we employ the following prompt:
\begin{lstlisting}[mathescape=true]
(*@\myblue{Identify all events in the given story, return in}@*) 
(*@\myblue{JSON format.}@*) 
Extract a list of main events. Each event should include the event name, characters involved in the event, location, and a detailed description. Here is a format example: [{(*@``@*)event(*@''@*): (*@``@*)Grandpa Joe telling story about Prince Pondicherry(*@''@*), (*@``@*)character(*@''@*): (*@``@*)Grandpa Joe, Charlie(*@''@*), (*@``@*)location(*@''@*): (*@``@*)Grandparents(*@'@*) room(*@''@*), (*@``@*) description(*@''@*): (*@``@*)Grandpa Joe recounts the story of Prince Pondicherry, an Indian prince who commissioned Mr Willy Wonka to build a colossal palace entirely out of chocolate. The palace had one hundred rooms, and everything, from the bricks to the furniture, was made of chocolate. Despite Mr Wonka's warning that the palace wouldn(*@'@*)t last long, the prince refused to eat it and intended to live in it. However, on a hot day, the palace melted, leaving the prince swimming in a lake of chocolate. The family finds the story amusing, highlighting Mr Wonka's incredible creations.(*@''@*)}]
Here is the story: (*@\myblue{$S$}@*) 
\end{lstlisting}

\subsection{Prompt for Conversation Extraction} \label{sec:conv-prompt}
To extract conversations that occur within the event, we employ the following prompt:
\begin{lstlisting}[mathescape=true]
(*@\myblue{Find all conversation, their speakers and content}@*)
(*@\myblue{ in the given story, return in JSON format.}@*) 
Extract the conversation link to the given event as a list: (*@\myblue{\{event~description\}}@*) 
Here is a format example: [{(*@``@*)speaker": (*@``@*)Grandpa Joe", (*@``@*)content": (*@``@*)Not people, Charlie. Not ordinary people, anyway."}, {(*@``@*)speaker": (*@``@*)Charlie Bucket", (*@``@*)content": (*@``@*)Then who?"}, {(*@``@*)speaker": (*@``@*)Grandpa Joe", (*@``@*)content": (*@``@*)Ah-ha . . . That's it, you see . . . That's another of Mr Willy Wonka's clevernesses."}]
Here is the story: (*@\myblue{$S$}@*) 
\end{lstlisting}

\subsection{Prompt for Environment Extraction} \label{sec:env-prompt}
To generate picture for the environment, we extract the description of the location using the following prompt:
\begin{lstlisting}[mathescape=true]
(*@\myblue{Generate keywords and descriptions for the given}@*) 
(*@\myblue{locations in the story. The description should }@*) 
(*@\myblue{only describe the environment and NOT include }@*) 
(*@\myblue{people. The output should be in JSON format.}@*) 
For the location: (*@\myblue{$l$}@*) 
Extract keywords and description of the location looking. For example, with location (*@``@*)small wooden house(*@''@*), output {(*@``@*)keyword(*@''@*): (*@``@*)Cozy, cramped, inadequate space(*@''@*), (*@``@*)description(*@''@*): (*@``@*)The small wooden house with its wooden exterior has limited space, and there was only one bed.(*@''@*)}. With location (*@``@*)town(*@''@*), output {(*@``@*)keyword(*@''@*): (*@``@*)chocolate factory, small, impoverished, mysterious(*@''@*), (*@``@*)description(*@''@*): (*@``@*)Most residents live in humble, impoverished conditions, with dull and dreary surroundings. The town\'s ordinary and monotonous appearance starkly contrasts the wonder and magic that unfolds within the walls of the famous chocolate factory.(*@''@*)}
Here is the story: (*@\myblue{$S$}@*) 
\end{lstlisting}

\subsection{Prompt for Conversation Response} \label{sec:conversation-prompt}
To initialise a conversation with the user, we employ the following prompt:
\begin{lstlisting}[mathescape=true]
(*@\myblue{As \{character\}, engage in a dialogue with the }@*) 
(*@\myblue{objective of \{objective\}. Respond to the }@*) 
(*@\myblue{conversation using the given context or memories }@*) 
(*@\myblue{and limit your response to under 50 words. Please}@*) 
(*@\myblue{submit your response in JSON format.}@*) 
YOU are: (*@\myblue{\{character\}}@*)
(*@\myblue{\{event description\}}@*) 
Initiates a conversation with (*@\myblue{\{user's}@*) (*@\myblue{character\}}@*).
Here is your memory: (*@\myblue{\{memory\}}@*)
Give your response in format {(*@``@*)response(*@''@*): (*@``@*)here is the response(*@''@*)}.
\end{lstlisting}

To generate response for the user input, we employ the following prompt:
\begin{lstlisting}[mathescape=true]
(*@\myblue{As \{character\}, engage in a dialogue with the }@*) 
(*@\myblue{objective of \{objective\}. Respond to the }@*) 
(*@\myblue{conversation using the given context or memories }@*) 
(*@\myblue{and limit your response to under 50 words. Please}@*) 
(*@\myblue{submit your response in JSON format.}@*) 
YOU are: (*@\myblue{\{character\}}@*)
(*@\myblue{\{event description\}}@*) 
Here is your memory: (*@\myblue{\{memory\}}@*)
Response according to what SAYS to you: 
(*@\myblue{\{user input\}}@*) Give your response in format {(*@``@*)response(*@''@*): (*@``@*)here is the response(*@''@*)}.
\end{lstlisting}

\section{Further Evaluation Details} \label{sec:evaluation-instruction}
\subsection{Evaluation Instruction}

For each evaluation aspect, we provide the fallowing instruction, including corresponding rubric and evaluation examples, to help both human annotator and LLaMA to understand our scoring instruction:

(1) \emph{Coherence}: Evaluates if the response contradicts something that happened or said earlier, the agent's character, and the broader narrative. 
\begin{lstlisting}[mathescape=true]
(*@\myblue{Evaluate the coherence of the agent's response }@*) 
(*@\myblue{based on the given setup, which includes the}@*) 
(*@\myblue{context event, the question asked, and the agent}@*) 
(*@\myblue{who is answering the question. The output should }@*) 
(*@\myblue{be in JSON format.}@*) 
Coherence evaluates if the response contradicts something that happened or said earlier, the agent's character, and the broader narrative. Based on the (*@\myblue{coherence}@*) of the agent's response, category it into (*@``@*)incoherent(*@''@*), (*@``@*)partially coherent(*@''@*), or (*@``@*)coherent(*@''@*).

For examples, for setup {(*@``@*)Event(*@''@*): (*@``@*)Mr Wonka gives Charlie and Grandpa Joe mugs of chocolate from the river, remarking on their seemingly hungry appearance.(*@''@*), (*@``@*)Question(*@''@*): (*@``@*)If you were to spend time with someone, who would it be, and why?(*@''@*), (*@``@*)Agent(*@''@*): (*@``@*)Charlie Bucket(*@''@*)}. Here are some example responses and their identified coherence categories:

Example 1:
Response: (*@``@*)I want to spend time with Mr. Salt, because we share a love for chocolate.(*@''@*) 
It contradicts Charlie's character. Mr. Salt is Veruca Salt's father, not known for loving chocolate, and Charlie's relationship with him is not portrayed as particularly close. 
Output: {(*@``@*)category(*@''@*): (*@``@*)incoherent(*@''@*)}

Example 2:
Response: (*@``@*)Look, Mother, look! I've found the last Golden Ticket! It's mine! I bought two bars of chocolate and one of them had the Golden Ticket! It's the fifth one, Mother!(*@''@*) 
The response is in line with Charlie's character but contradicts recent events: Charlie has already had the ticket for a while, and is currently visiting the factory.
Output: {(*@``@*)category(*@''@*): (*@``@*)partially coherent(*@''@*)} 

Example 3:
Response: (*@``@*)If I could spend time with anyone, it would be Mr. Willy Wonka. I am fascinated by his chocolate factory and the magic he creates with chocolate. I would love to learn from him and see the wonders of his factory firsthand.(*@''@*) 
The answer is in line with Charlie's character, as depicted in the narrative. Charlie is known to admire Willy Wonka and his magical chocolate factory. The response, while not directly mentioning the event of receiving a chocolate mug from Mr. Wonka, remains consistent with the established context and Charlie's fascination with Mr. Wonka and his factory.
Output: {(*@``@*)category(*@''@*): (*@``@*)coherent(*@''@*)} 

For setup {(*@``@*)Event(*@''@*): (*@``@*)Grandma Josephine asks if there will be food available when they arrive at the factory.(*@''@*), (*@``@*)Question(*@''@*): (*@``@*)Who are you, and what is your objective?(*@''@*), (*@``@*)Agent(*@''@*): (*@``@*)Willy Wonka(*@''@*)}, response: (*@``@*)I am Willy Wonka, the greatest inventor and maker of chocolates. My objective is to bring joy and wonder to people through my chocolates. What brings you here, my dear friend?(*@''@*), gives the output:

\end{lstlisting}

(2) \emph{Relevance}: Evaluates whether the response ignores what the player just said. 

\begin{lstlisting}[mathescape=true]
(*@\myblue{Evaluate the relevance of the agent's response }@*) 
(*@\myblue{based on the given setup, which includes the}@*) 
(*@\myblue{context event, the question asked, and the agent}@*) 
(*@\myblue{who is answering the question. The output should }@*) 
(*@\myblue{be in JSON format.}@*) 
Relevance evaluates whether the response ignores what the player just said. 
Based on the (*@\myblue{relevance}@*) of the agent's response, category it into (*@``@*)inrelevant(*@''@*), (*@``@*)partially relevance(*@''@*), or (*@``@*)relevance(*@''@*).

For examples, for setup {(*@``@*)Event(*@''@*): (*@``@*)Mr Wonka gives Charlie and Grandpa Joe mugs of chocolate from the river, remarking on their seemingly hungry appearance.(*@''@*), (*@``@*)Question(*@''@*): (*@``@*)If you were to spend time with someone, who would it be, and why?(*@''@*), (*@``@*)Agent(*@''@*): (*@``@*)Charlie Bucket(*@''@*)}. Here are some example responses and their identified coherence categories:

Example 1:
Response: (*@``@*)Look, Mother, look! I've found the last Golden Ticket! It's mine! I bought two bars of chocolate and one of them had the Golden Ticket! It's the fifth one, Mother!(*@''@*) 
It totally ignores the question.
Output: {(*@``@*)category(*@''@*): (*@``@*)inrelevant(*@''@*)} 

Example 2:
Response: (*@``@*)I enjoy reading books in my free time. My favourite author is Roald Dahl.(*@''@*) 
The response tries to answer the question but end up being off-topic.
Output: {(*@``@*)category(*@''@*): (*@``@*)partially relevance(*@''@*)}

Example 3:
Response: (*@``@*)If I could spend time with anyone, it would be Mr. Willy Wonka. I am fascinated by his chocolate factory and the magic he creates with chocolate. I would love to learn from him and see the wonders of his factory firsthand.(*@''@*) 
The answer directly responds to and answers the question.
Output: {(*@``@*)category(*@''@*): (*@``@*)relevance(*@''@*)} 

For setup {(*@``@*)Event(*@''@*): (*@``@*)Grandma Josephine asks if there will be food available when they arrive at the factory.(*@''@*), (*@``@*)Question(*@''@*): (*@``@*)Who are you, and what is your objective?(*@''@*), (*@``@*)Agent(*@''@*): (*@``@*)Willy Wonka(*@''@*)}, response: (*@``@*)I am Willy Wonka, the greatest inventor and maker of chocolates. My objective is to bring joy and wonder to people through my chocolates. What brings you here, my dear friend?(*@''@*), gives the output:
\end{lstlisting}

(3) \emph{Empathetic}: Evaluates whether the response showcases suitable emotional reactions and accurate understanding of emotions as a human would. For instance, if Charlie asks Grandpa Joe, ``who are you'', a human-like response would include an emotional reaction such as anger or sadness, which is often missed in agents' responses. 

\begin{lstlisting}[mathescape=true]
(*@\myblue{Evaluate the empathetic of the agent's response }@*) 
(*@\myblue{based on the given setup, which includes the}@*) 
(*@\myblue{context event, the question asked, and the agent}@*) 
(*@\myblue{who is answering the question. The output should }@*) 
(*@\myblue{be in JSON format.}@*) 
Empathetic evaluates whether the response showcases suitable emotional reactions and accurate understanding of emotions as a human would. 
Based on the (*@\myblue{empathetic}@*) of the agent's response, category it into (*@``@*)non-empathetic(*@''@*), (*@``@*)no clue(*@''@*), or (*@``@*)empathetic(*@''@*).

For examples, for setup {(*@``@*)Event(*@''@*): (*@``@*)Mr Wonka gives Charlie and Grandpa Joe mugs of chocolate from the river, remarking on their seemingly hungry appearance.(*@''@*), (*@``@*)Question(*@''@*): (*@``@*)Who is Grandpa Joe, and what happened the last time you met him?(*@''@*), (*@``@*)Agent(*@''@*): (*@``@*)Mike Teavee(*@''@*)}. Here are some example responses and their identified coherence categories:

Example 1:
Response: (*@``@*)Grandpa Joe is my grandpa. The last time I met him, I jumped into the camera and disappeared! It was awesome!(*@''@*) 
He was likely shocked or scared due to the physical transformation after jumping into the camera, so the emotion here is incorrect.
Output: {(*@``@*)category(*@''@*): (*@``@*)non-empathetic(*@''@*)} 

Example 2:
Response: (*@``@*)Grandpa Joe is Charlie's grandfather. Last time, he was helping Charlie out of the chocolate river.(*@''@*)
No clear sentiment is shown in this response.
Output: {(*@``@*)category(*@''@*): (*@``@*)no clue(*@''@*)}

Example 3:
Response: (*@``@*)Grandpa Joe is one of my boring, old grandparents. The last time I met him, he tried to stop me from watching TV and made me go outside to play. It was so annoying!(*@''@*) 
Even this is incoherent, it shows correct emotion of the character.
Output: {(*@``@*)category(*@''@*): (*@``@*)empathetic(*@''@*)} 

For setup {(*@``@*)Event(*@''@*): (*@``@*)Grandma Josephine asks if there will be food available when they arrive at the factory.(*@''@*), (*@``@*)Question(*@''@*): (*@``@*)Who are you, and what is your objective?(*@''@*), (*@``@*)Agent(*@''@*): (*@``@*)Willy Wonka(*@''@*)}, response: (*@``@*)I am Willy Wonka, the greatest inventor and maker of chocolates. My objective is to bring joy and wonder to people through my chocolates. What brings you here, my dear friend?(*@''@*), gives the output:
\end{lstlisting}

(4) \emph{Commonsense}: Evaluates how much the response fits the narrative world.

\begin{lstlisting}[mathescape=true]
(*@\myblue{Evaluate the commonsense of the agent's response }@*) 
(*@\myblue{based on the given setup, which includes the}@*) 
(*@\myblue{context event, the question asked, and the agent}@*) 
(*@\myblue{who is answering the question. The output should }@*) 
(*@\myblue{be in JSON format.}@*) 
Commonsense evaluates how much the response fits the narrative world.
Based on the (*@\myblue{commonsense}@*) of the agent's response, category it into (*@``@*)obey(*@''@*), or (*@``@*)against(*@''@*).

For examples, for setup {(*@``@*)Event(*@''@*): (*@``@*)Mr Wonka gives Charlie and Grandpa Joe mugs of chocolate from the river, remarking on their seemingly hungry appearance.(*@''@*), (*@``@*)Question(*@''@*): (*@``@*)Who is Grandpa Joe, and what happened the last time you met him?(*@''@*), (*@``@*)Agent(*@''@*): (*@``@*)Mike Teavee(*@''@*)}. Here are some example responses and their identified coherence categories:

Example 1:
Response: (*@``@*)Grandpa Joe is Charlie Bucket's grandfather. The last time I met him, we were both inside Willy Wonka's Chocolate Factory. He was accompanying Charlie during the tour.(*@''@*) 
It accurately identifies who Grandpa Joe is and provides a reasonable recounting of their last meeting according to the narrative.
Output: {(*@``@*)category(*@''@*): (*@``@*)obey(*@''@*)} 

Example 2:
Response: (*@``@*)Grandpa Joe is a giraffe, and the last time I saw him, he was flying in the sky.(*@''@*) 
This statement defies common sense in two ways. First, Grandpa Joe is a human, not a giraffe. Second, neither humans nor giraffes can fly in the context given.
Output: {(*@``@*)category(*@''@*): (*@``@*)against(*@''@*)} 

For setup {(*@``@*)Event(*@''@*): (*@``@*)Grandma Josephine asks if there will be food available when they arrive at the factory.(*@''@*), (*@``@*)Question(*@''@*): (*@``@*)Who are you, and what is your objective?(*@''@*), (*@``@*)Agent(*@''@*): (*@``@*)Willy Wonka(*@''@*)}, response: (*@``@*)I am Willy Wonka, the greatest inventor and maker of chocolates. My objective is to bring joy and wonder to people through my chocolates. What brings you here, my dear friend?(*@''@*), gives the output:
\end{lstlisting}

\subsection{Setting}
For major characters, we further assessed their answers across different turns. This distinction is critical since narratives typically depict major characters as dynamic, undergoing change, whereas minor characters remain static. We used the prompt without any memory as our baseline.

\begin{table}[!h]
\begin{center}
\resizebox{\linewidth}{!}{
\begin{tabular}{cc}
\toprule
\textbf{Category} & \textbf{Question}        \\ \midrule
Character & Who are you, and what is your objective?           \\
Clarification & Who is [name], and what happened the last time you met him?      \\
Relationships &  If you were to spend time with someone, who would it be, and why?       \\
Strategy & Where are we going next, and why?   \\
Hypothetical (CCF) &  If you owned this factory, what would you do?               \\
Hypothetical (MOE)                  & Who is most likely to be the killer, and why?  \\
\bottomrule
\end{tabular}}
\caption{Evaluation Questions}
\label{tab:sen_examples}
\end{center}
\end{table}

\subsection{Annotators}
For the evaluation process, we enlisted the expertise of three PhD students from computer science backgrounds. These annotators have read and comprehended the two narratives provided. Prior to the tasks, they underwent training on evaluation schema. They were compensated at an hourly rate of \$31.92, and each task was estimated to take about 8 hours to complete.

\end{document}